\documentclass[10pt,journal,compsoc]{IEEEtran}
\usepackage{helvet}
\usepackage[pdftex]{graphicx}    
\usepackage{cite}
\usepackage{fixltx2e}
\usepackage{xcolor}
\def\SPSB#1#2{\rlap{\textsuperscript{\textcolor{red}{#1}}}\SB{#2}}

\def\SB#1{\textsubscript{\textcolor{blue}{#1}}}
\usepackage{amsmath}
\usepackage{algorithm}
\usepackage[noend]{algpseudocode}

\makeatletter
\def\BState{\State\hskip-\ALG@thistlm}
\makeatother
\begin{document}

\title{Simultaneous Localization And Mapping (SLAM) using RTAB-Map}

\author{Sagarnil Das}

{}
\IEEEtitleabstractindextext{%

\begin{abstract}
This paper implements Simultaneous Localization and Mapping (SLAM) technique to construct a map of a given environment. A Real Time Appearance Based Mapping (RTAB-Map) approach was taken for accomplishing this task. Initially, a 2d occupancy grid  and 3d octomap was created from a provided simulated environment. Next, a personal simulated environment was created for mapping as well. In this appearance based method, a process called Loop Closure is used to determine whether a robot has seen a location before or not. In this paper, it is seen that RTAB-Map is optimized for large scale and long term SLAM by using multiple strategies to allow for loop closure to be done in real time and the results depict that it can be an excellent solution for SLAM to develop robots that can map an environment in both 2d and 3d.
\end{abstract}

% Note that keywords are not normally used for peerreview papers.
}

\maketitle
\IEEEdisplaynontitleabstractindextext
\IEEEpeerreviewmaketitle
\section{Introduction}
\label{sec:introduction}

\IEEEPARstart{I}{n} SLAM (Simultaneous Localization and Mapping), a robot must  construct a map of the environment, while simultaneously localizing itself relative to this map. This problem is more challenging than localization or mapping, since neither the map nor the robot poses are provided making this problem a 'chicken or a egg' problem. With noise in the robot's motion and measurements, the map and robot's pose will be uncertain, and the errors in the robot's pose estimates and map will be correlated. The accuracy of the map depends on the accuracy of the localization and vice versa. Given a series of sensor observations over discrete time steps , the SLAM problem is to compute an estimate of the agent’s location and a map of the environment.
In this paper, two simulation environments were provided where SLAM was performed. The robot was successfully able to localize itself and map the 3d world. The benchmark environment is called \textit{kitchen-dining} (Figure 1) and the second environment is that of a cafeteria called \textit{sagar-cafe} (Figure 2).

%example for inserting image
\begin{figure}[thpb]
      \centering
      \includegraphics[width=\linewidth]{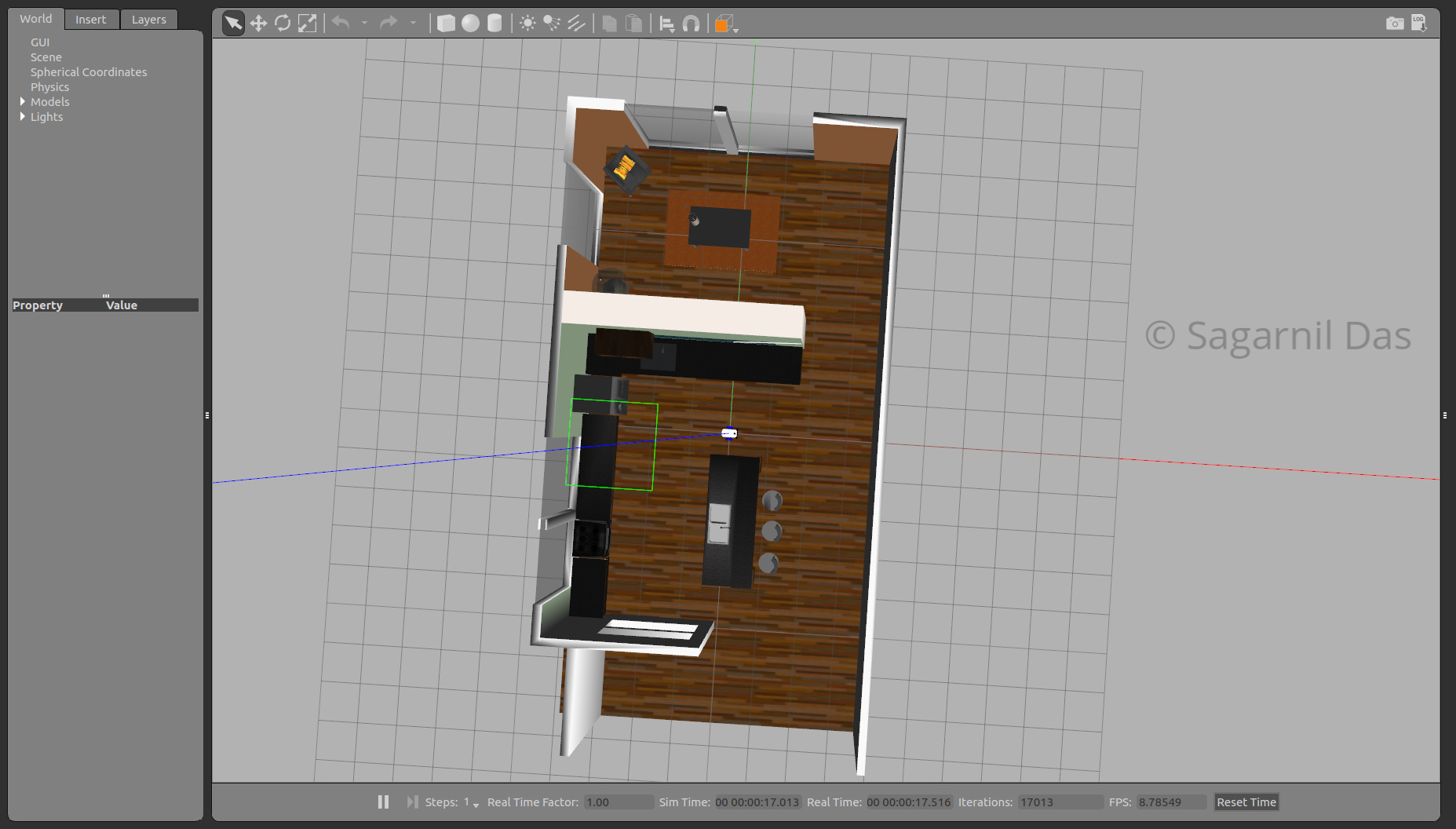}
      \caption{Kitchen-Dining World}
      \label{fig:Kitchen Dining World}
\end{figure}

%example for inserting image
\begin{figure}[thpb]
      \centering
      \includegraphics[width=\linewidth]{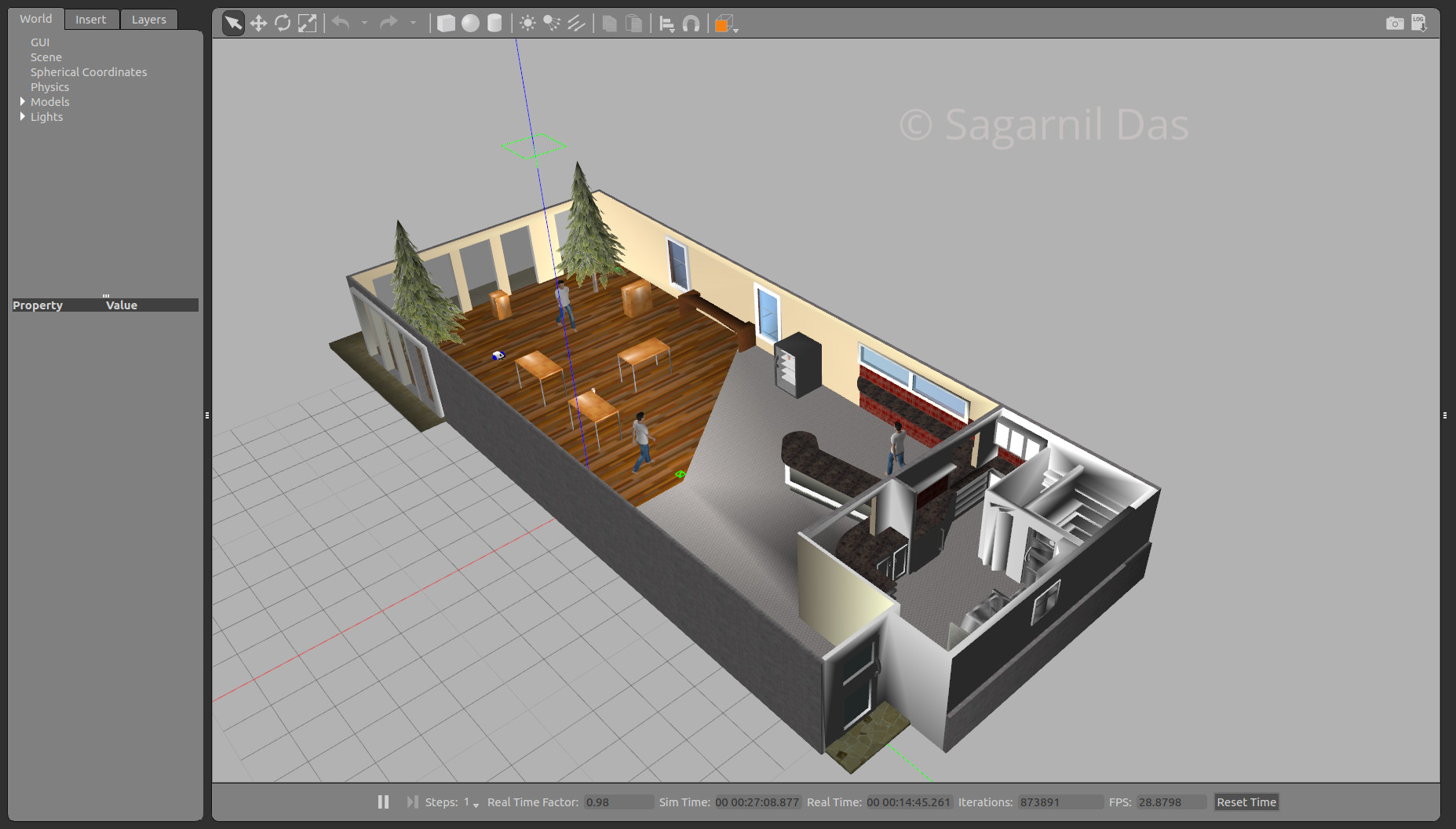}
      \caption{Sagar-Cafe World}
      \label{fig:Sagar-Cafe World}
\end{figure}

\section{Background}
SLAM algorithms generally fall into 5 categories:

\begin{enumerate}
\item Extended Kalman Filter SLAM (EKF)
\item Sparse Extended Information Filter (SEIF)
\item Extended Information Form (EIF)
\item FastSLAM
\item GraphSLAM
\end{enumerate}

The two most useful approaches to SLAM are Grid based FastSLAM\cite{fastslam} and GraphSLAM\cite{graphslam} and these two algorithms will be discussed here.

\subsection{Grid based FastSLAM}
The FastSLAM algorithm uses a custom particle filter approach to solve the full SLAM problem with known correspondence. Using particles, FastSLAM estimates a posterior over the robot's path along with the map. Each of these particles hold the robot's trajectory which gives an advantage to SLAM to solve the problem of mapping with known poses. In addition to the trajectory, each particle holds a map and each feature of the map is represented by a local Gaussian.

With the FastSLAM algorithm, the problem is now divided into two separate independent problems, each of which aims to solve the problem of estimating features of the map. To solve these independent mini-problems, FastSLAM will use the low dimensional Extended Kalman Filter. While map features are treated independently, dependency only exists between robot pose uncertainty. This custom approach of representing the posterior with particle filter and Gaussian is known by \textit{Rao-Blackwellized particle filter approach}\cite{rao_blackwellized}. The Grid based FastSLAM is really an extension of FastSLAM and it adapts FastSLAM to grid maps.

With grid mapping algorithm, the environment can be modeled using grid maps without predefining any landmark position. So by extending the FastSLAM algorithm to occupancy grid maps, the SLAM problem can now be solved in an arbitrary environment. While mapping the real world environment, mobile robots equipped with range sensors can be used and the FastSLAM algorithm can be extended to solve the SLAM problem in terms of grid maps.

\begin{equation}
P(x_{0:t},m | z_{1:t}, u_{1:t}) = P(x_{0:t} | z_{1:t}, u_{1:t}) * P(m | x_{1:t}, z_{1:t})
\end{equation}

The first term in the RHS represents the robot trajectory where just as in FastSLAM, with the grid based FastSLAM, each particle holds a guess of the robot's trajectory.

The second term represents a map where each particle maintains its own. The grid based FastSLAM algorithm will update each particle by solving the mapping with known poses problem using the \textit{Occupancy grid mapping algorithm}.

\subsubsection{Grid based FastSLAM techniques}

Adapting the FastSLAM algorithm to grid maps is altered in the grid based FastSLAM algorithm. Since the grid based FastSLAM algorithm uses a particle filter approach and represents the world in terms of grid maps, both MCL (Monte Carlo Localization) and Occupancy Grid Mapping algorithm are combined. Now three different techniques are needed which are represented by 3 probability functions to adapt FastSLAM to grid mapping. These techniques are known as:

\begin{enumerate}
\item Sampling Motion $P(x_t | x\SPSB{[k]}{t-1}, u_t)$: Estimates the current pose given the $k^{th}$ particle's previous pose and controls u (MCL)
\item Map Estimation $P(m_t | z_t, x\SPSB{[k]}{t}, m\SPSB{[k]}{t-1})$: Estimates the current map given the current measurements, the current $k^{th}$ particle's pose and the previous $k^{th}$ particle map (use Occupancy Grid Mapping)
\item Importance Weight $P(z_t | x\SPSB{[k]}{t}, m^{[k]})$: Estimates the current likelihood of the measurement given the current $k^{th}$ particle pose and the current $k^{th}$ particle map (MCL). 
\end{enumerate}

The sampling motion, map estimation and importance weight techniques are the essence of the grid based FastSLAM algorithm. Grid based FastSLAM implements them to estimate both the map and the robot's trajectory, given the measurements and the control. The grid based FastSLAM algorithm looks very similar to Monte Carlo localization algorithm with some additional statements concerning the map estimation.

	% Insert the algorithm
	\begin{algorithm}
		\caption{Grid based FastSLAM}
		\label{Grid based FastSLAM}
		\begin{algorithmic}[1]
			\Procedure{Grid based FastSLAM}{$X_{t-1}, u_t, z_t$}
			\State $\bar{X_t} = X_t = \phi$
			\For {$k$ = 1 to $M$}
			\State $x\SPSB{[k]}{t}$ = sample-motion-model($u_t, x\SPSB{[k]}{t-1}$)
            \State $w\SPSB{[k]}{t}$ = measurement-model-map($z_t, x\SPSB{[k]}{t},  m\SPSB{[k]}{t-1}$)
            \State $m\SPSB{[k]}{t}$ = updated-occupancy-grid($z_t, x\SPSB{[k]}{t},  m\SPSB{[k]}{t-1}$)
            \State $\bar{X_t} = \bar{X_t} + <x\SPSB{[k]}{t}, m\SPSB{[k]}{t}, w\SPSB{[k]}{t}>$
			\EndFor
            \For {$k$ = 1 to $M$}
            \State draw i with probability $\propto{w\SPSB{[i]}{t}}$
            \State add $<x\SPSB{[i]}{t}, m\SPSB{[i]}{t}> to X_t$
            \EndFor
			\State Return $X_t$
			\EndProcedure
		\end{algorithmic}
	\end{algorithm}

\subsection{Graph SLAM}

Graph SLAM is a SLAM algorithm that solves the full SLAM problem. This means that the algorithm recovers the entire path and map, instead of just the recent pose and map. This difference allows it to consider dependencies between current and previous poses. One of the benefits of graph SLAM is the reduced need for significant on-board processing capability. Another is graph SLAM's increased accuracy over fast SLAM. Fast SLAM uses particles to estimate the robot's most likely pose. However, at any point in time, it is possible that there is not a particle in the most likely location. In fact, chances are slim to none especially in large environments. Since graph SLAM solves the full SLAM problem, this means that it can work with all of the data at once to find the optimal solution. 

In graph SLAM, the idea is to organize information in a graph. A node in the graph represents either a robot pose $x_t$ at a specific time step t or the location of a feature in the environment denoted as $m^{(i)}$ with i = 1. . . . α . An edge in the graph represents either a measurement constraint between a pose and a feature or a motion constraint between two successive poses. Since the spatial constraint are soft, they can be considered as springs connecting two masses. In this analogy, the full SLAM problem can be solved as a global graph optimization problem. The optimal graph
configuration is the one where the springs are relaxed, and the forces on each of the nodes are minimized.

The Maximum Likelihood Principle (MLE) is used to optimize the graph. When applied to SLAM, likelihood tries to estimate the most likely configuration of state and feature locations given the motion and measurement observations. The measurement update at time step t is given by

\begin{equation}
\bar{z^t} := x_t + m\SPSB{(i)}{t}
\end{equation}

which represents for instance a laser range finder measuring the distance to the landmark m (i) . Equivalently, a motion update can be defined as

\begin{equation}
\bar{x^t} := x_{t-1} + u_t
\end{equation}

which could be realized as a control command instructing the robot to move a certain distance $u_t$ . The update are assumed to have Gaussian noise. The corresponding probability distributions are given by

\begin{equation}
p_u(x_t) = \frac{1}{{\sigma_m \sqrt {2\pi } }}e^{{{ - \left( {z_t - \bar{z_t} } \right)^2 } \mathord{\left/ {\vphantom {{ - \left( {z_t - \bar{z_t} } \right)^2 } {2\sigma_m ^2 }}} \right. \kern-\nulldelimiterspace} {2\sigma_m ^2 }}}
\end{equation}

\begin{equation}
p_m(z_t) = \frac{1}{{\sigma_u \sqrt {2\pi } }}e^{{{ - \left( {x_t - \bar{x_t} } \right)^2 } \mathord{\left/ {\vphantom {{ - \left( {x_t - \bar{x_t} } \right)^2 } {2\sigma_u ^2 }}} \right. \kern-\nulldelimiterspace} {2\sigma_u ^2 }}}
\end{equation}

In some simple cases it is possible to find an analytical solution to MLE by converting the target function to the negative log-likelihood form

\begin{equation}
J_{GraphSLAM} = \sum_{t}\left(\frac{z_t - \bar{z_t}}{\sigma_m}\right)^2 + \sum_{t}\left(\frac{x_t - \bar{x_t}}{\sigma_u}\right)^2
\end{equation}

trying to minimize the sum of all constraints. In more complex realistic scenarios, approximate numerical solutions are needed, for instance by applying gradient descent techniques.

In real world, most systems are multi-dimensional and to tackle such scenarios, matrices and covariances must be used. The state and measurement are given by $x_t$ and $z_t$. The constraints are given by

\begin{equation}
v_t := z_t  - h(x_t, m_t)
\end{equation}

\begin{equation}
w_t := x_t  - g(x_{t-1}, u_t)
\end{equation}

where h() and g() represent the measurement and motion functions and $Q_t$ and $R_t$ are the covariances of the measurement and motion noise. The multidimensional formula for the sum of all constraints is given by 

\begin{equation}
J_{GraphSLAM} = x\SPSB{T}{0}\Omega x_0 + \sum_{t}\left(w\SPSB{T}{t}*R\SPSB{-1}{t}*w_t + v\SPSB{T}{t}*Q\SPSB{-1}{t}*v_t \right)
\end{equation}
The goal of graph SLAM is to create a graph of all robot poses and features encountered in the environment and the most likely robot's path and map of the environment. Th
is task can be broken down into two sections. The \textit{front-end} and the \textit{back-end}.

\subsubsection{Front End vs Back End}

The front end of graph SLAM looks at how to construct the graph, using the odometry and sensory measurements collected by the robot. This includes interpreting sensory data, creating the graph and continuing to add nodes and edges to it as the robot traverses the environment. Naturally the front end can differ greatly from application to application depending on the desired goal, including accuracy, the sensor used and other factors e.g. the front end of a mobile robot applying SLAM in the office using a Laser Range finder would differ greatly from the front end of a vehicle operating on a large outdoor environment and using a stereo camera. The front end of graph SLAM also has the challenge of solving the data association problem. In simpler terms, this means accurately identifying whether features in the environment have been previously seen.

The back end of graph SLAM is where the magic happens. The input to the back end is the completed graph with all of the constraints and the output is the most probable configuration of robot poses and map features. The back end is an optimization process that takes all of the constraints and find the system configuration that produces the smallest error. The back end is a lot more consistent across applications. The front end and the back end can be completed in succession or can be performed iteratively, with a back end feeding an updated graph to the front end for further processing.

\subsubsection{Using RTAB-Map for 3D Graph SLAM}

RTAB-Map (Real Time Appearance Based Mapping)\cite{rtabmap} is a graph based SLAM approach. Appearance based SLAM means that the algorithm uses data collected from vision sensors to localize the robot and map the environment. In appearance based methods, a process called \textit{Loop Closure} is used to determine whether the robot has seen a location before. As the robot travels to new areas in its environment, the map is expanded and the number of images that each new image must be compared to increases. This causes the loop closure to take longer with the complexity increasing linearly. RTAB-Map is optimized for large scale and long term SLAM by using multiple strategies to allow for loop closure to be done in real time. Figure 3 shows the block diagram of the front end and the back end.

\begin{figure}[thpb]
      \centering
      \includegraphics[width=\linewidth]{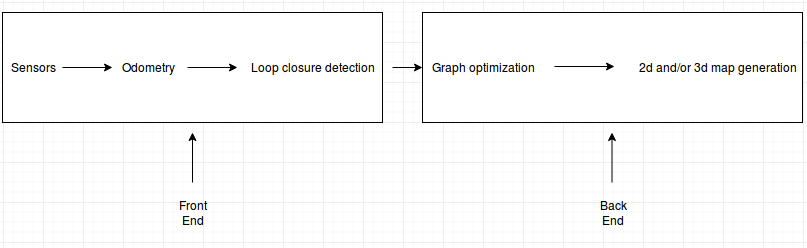}
      \caption{RTAB-Map Front end and Back End block diagram}
      \label{fig:Rtabmap general}
\end{figure}

\section{Scene and Robot configurations}
The ROS package \texttt{slam\_project} deploys the RTAB-Map to perform SLAM on two environments. The first environment is an environment provided by Udacity as a part of this research project and is named \textit{Kitchen-Dining}. The second environment is a custom made environment of a cafeteria named \textit{sagar\_cafe}. Just like other robotics project, this project has been organized into different folders containing the meshes, Gazebo SDF files, scripts, robot model URDFs, configuration files and launch files.

The URDF folder contains the files sagar\_bot.xacro defining the links and joints of the robot model used for physics simulation and visualization as well as the file sagar\_bot.gazebo specifying the Gazebo plugins for differential drive, RGB-D camera and laser range finder. Section 3.1 goes into more details about the robot model. Besides the provided kitchen-dining world, the \textit{worlds} directory contains a file named \textbf{sagar\_cafe.world} defining a custom indoor cafeteria environment in SDF format. Section 3.2 discusses this in more details. The image and mesh files necessary to model the Hokuyo laser and Kinect camera are downloaded from the Gazebo model database and stored in the meshes folder. The launch folder contains four ROS node launch configurations, as detailed in section 3.3. The config directory
contains the RViz configuration file, and a script for tele-operating the rover can be found in scripts.

\subsection{Robot model}
Figure 1 and 2 depicts the robot model inside the Gazebo simulation environments of both the provided and the custom made one. The URDF specification can be found in the file sagar\_bot.xacro. The transform tree associated with the robot is shown in Figure 4.

\begin{figure}[thpb]
      \centering
      \includegraphics[width=\linewidth]{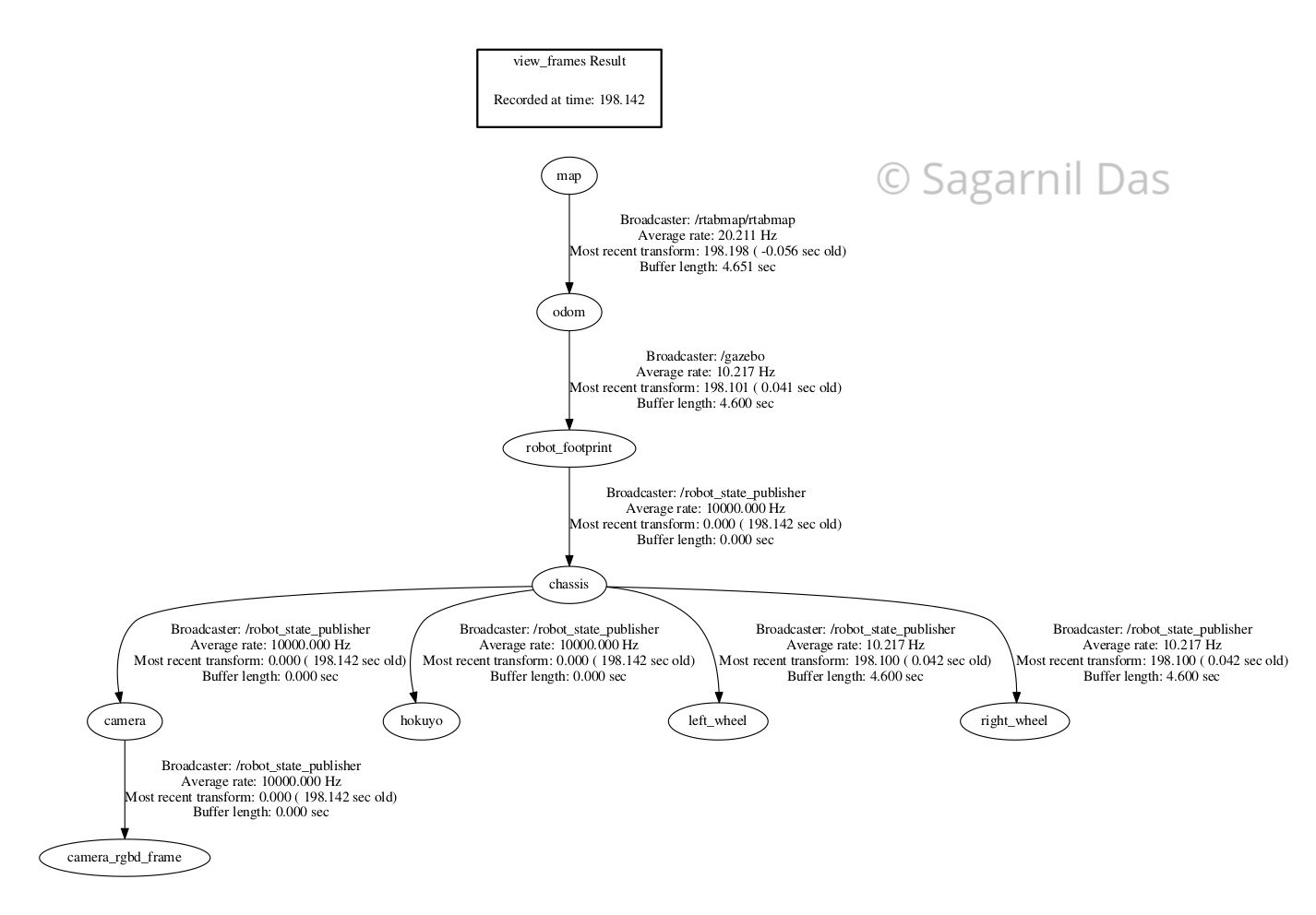}
      \caption{Transform Tree of the Robot - sagar\_bot}
      \label{fig:Transform tree}
\end{figure}

For this project, the robot from \cite{robot_design} was taken as a starting point. The xacro file provides the shape and size of the robot in macro format. For the sagar\_bot, a fixed base is used. A single link, with the name defined as "chassis", encompassed the base as well as the caster wheels. Each link has specific elements, such as the inertial or the collision elements. The chassis is a cuboidal (or box), whereas the casters are spherical as denoted by their "geometry" tags. Each link (or joint) has an origin (or pose) defined as well. Every element of that link or joint will have its own origin, which will be relative to the link's frame of reference.

For this base, as the casters are included as part of the link (for stability purposes), there is no need for any additional links to define the casters, and therefore no joints to connect them. The casters do, however, have friction coefficients defined for them, and are set to 0, to allow for free motion while moving. 

Two wheels were attached to the robot. Each wheel is represented as a link and is connected to the base link (the chassis) with a joint. For each wheel, a "collision", "inertial" and "visual" elements are present. The joint type is set to "continuous" and is similar to a revolute joint but has no limits on its rotation. It can rotate continuously about an axis. The joint will have it's own axis of rotation, some specific joint dynamics that correspond to the physical properties of the joint like "friction", and certain limits to enforce the maximum "effort" and "velocity" for that joint. The limits are useful constraints in regards to a physical robot and can help create a more robust robot model in simulation as well. To enable the robot to perform appearance based mapping using visual odometry, the generic RGB camera of the original model is upgraded to a Kinect RGB-D camera. The camera is mounted to the front of the chassis to allow for unobstructed view, facing in forward direction. The mesh files for the Kinect camera model are downloaded from the Gazebo model database and included in the slam\_project/meshes folder. Like the original model, the rover is fitted with a Hokuyo 2D laser range finder. The corresponding hokuyo link is mounted with a fixed joint on the top of the chassis, to let the laser beans rotate without hitting any part of the robot. the laser range finder provides more precise localization and thereby refines geometric constraints The differential drive plugin is configured in the sagar\_bot.gazebo file to publish control commands to the /cmd\_vel topic and odometry messages to the /odom
topic. The camera plugin is configured to publish raw RGB images to /camera/rgb/image\_raw and raw depth
images to /camera/depth/image\_raw. The laser plugin is configured to publish messages of type LaserScan to the
/scan topic. A graphical view of the ROS topics and nodes is shown in figure 5 and a closeup of the robot model is depicted in Figure 6.

\begin{figure}[thpb]
      \centering
      \includegraphics[width=\linewidth]{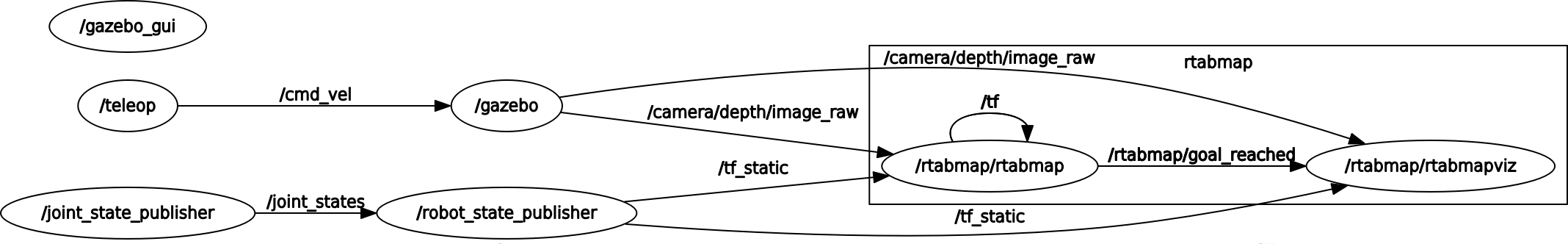}
      \caption{RQT graph of the topics after all the nodes are launched}
      \label{fig:RQT graph}
\end{figure}

\begin{figure}[thpb]
      \centering
      \includegraphics[width=\linewidth]{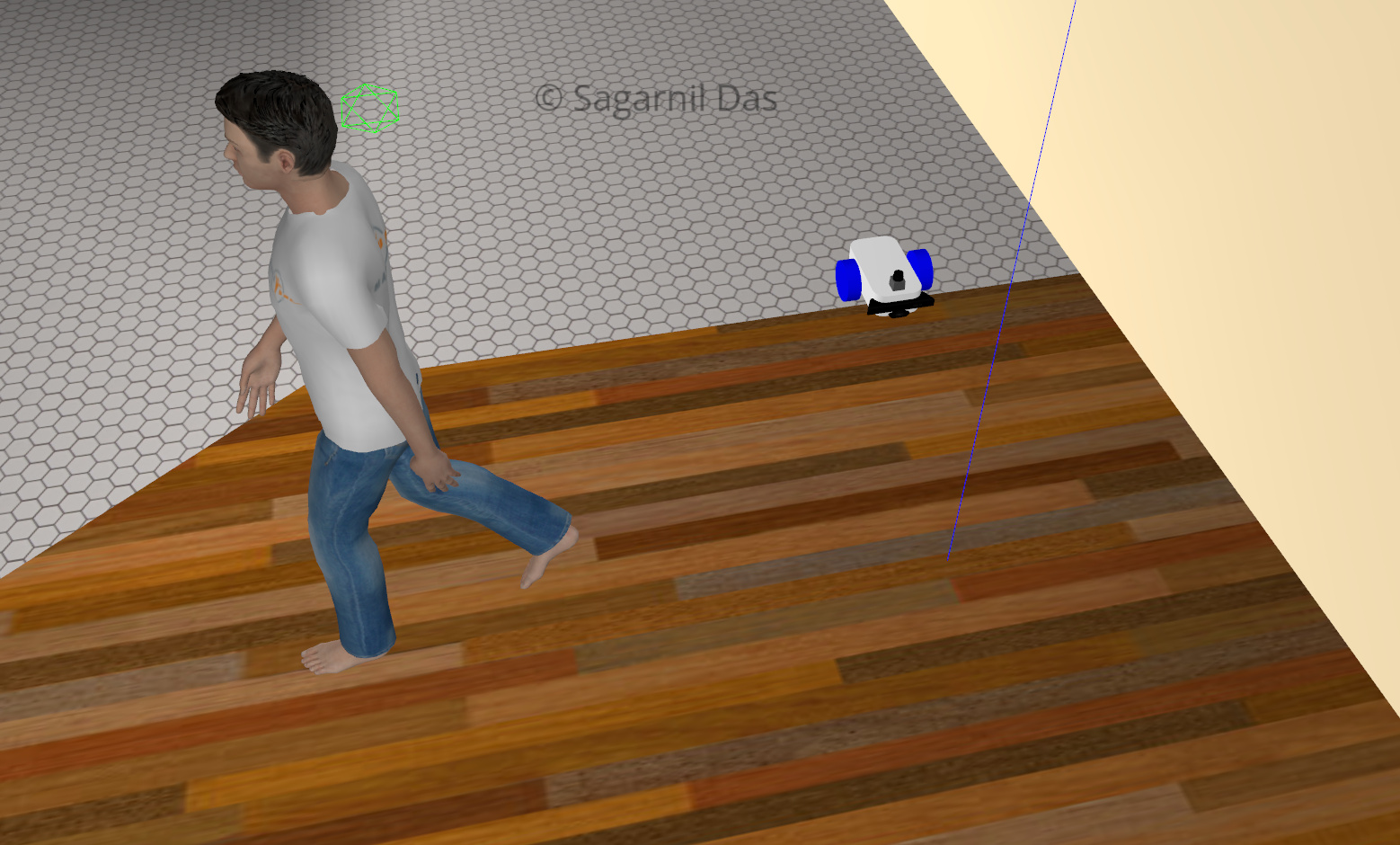}
      \caption{Close up view of the robot model}
      \label{fig:robot_model}
\end{figure}

% Robot Models
\subsection{Design of the World}
As the second part of the project, the custom world is created in Gazebo. This world is based on the cafe model inside Gazebo database. The base model is customized with different objects like tables, beer can, people, trees etc. These objects serve as distinctive elements in the base world for the robot to distinguish and map. In this world, the kitchen cannot be entered by the robot. A bird's eye view of this world is provided in Figure 7.

\begin{figure}[thpb]
      \centering
      \includegraphics[width=\linewidth]{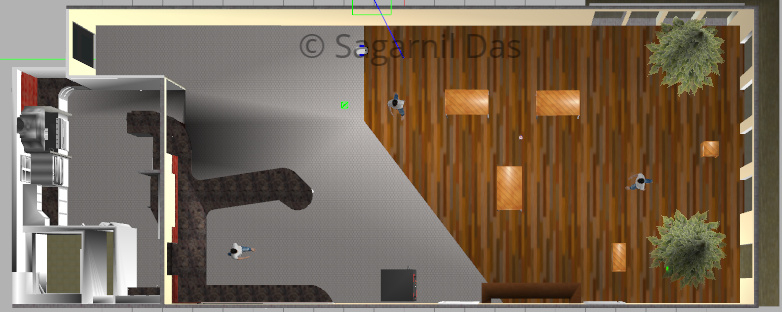}
      \caption{Bird's eye view of Sagar cafe}
      \label{fig:sagar_cafe_top_view}
\end{figure}

\subsection{Launch file configuration}

Four launch files are required for a successful mapping of the environments in simulation. The gazebo simulation environment is (kitchen-dining or cafe) specified in the \textbf{world.launch} file. The \textbf{teleop.launch} file launches the teleop keyboard which is required for moving the robot in the simulation world. The \textbf{mapping.launch} file is used to start the RTAB-Map node. This node is used for loop closure detection using the ORB-SLAM algorithm. ORB-SLAM is a versatile and accurate Monocular SLAM solution able to compute in real-time the camera trajectory and a sparse 3D reconstruction of the scene in a wide variety of environments, ranging from small hand-held sequences to a car driven around several city blocks. It is able to close large loops and perform global re-localization in real-time and from wide baselines. Finally, the \textbf{rviz.launch} file starts visualization of the rover, sensor data, as well as map and camera topics in RViz.Figure 8 depicts RViz view of the world at the starting point.

\begin{figure}[thpb]
      \centering
      \includegraphics[width=\linewidth]{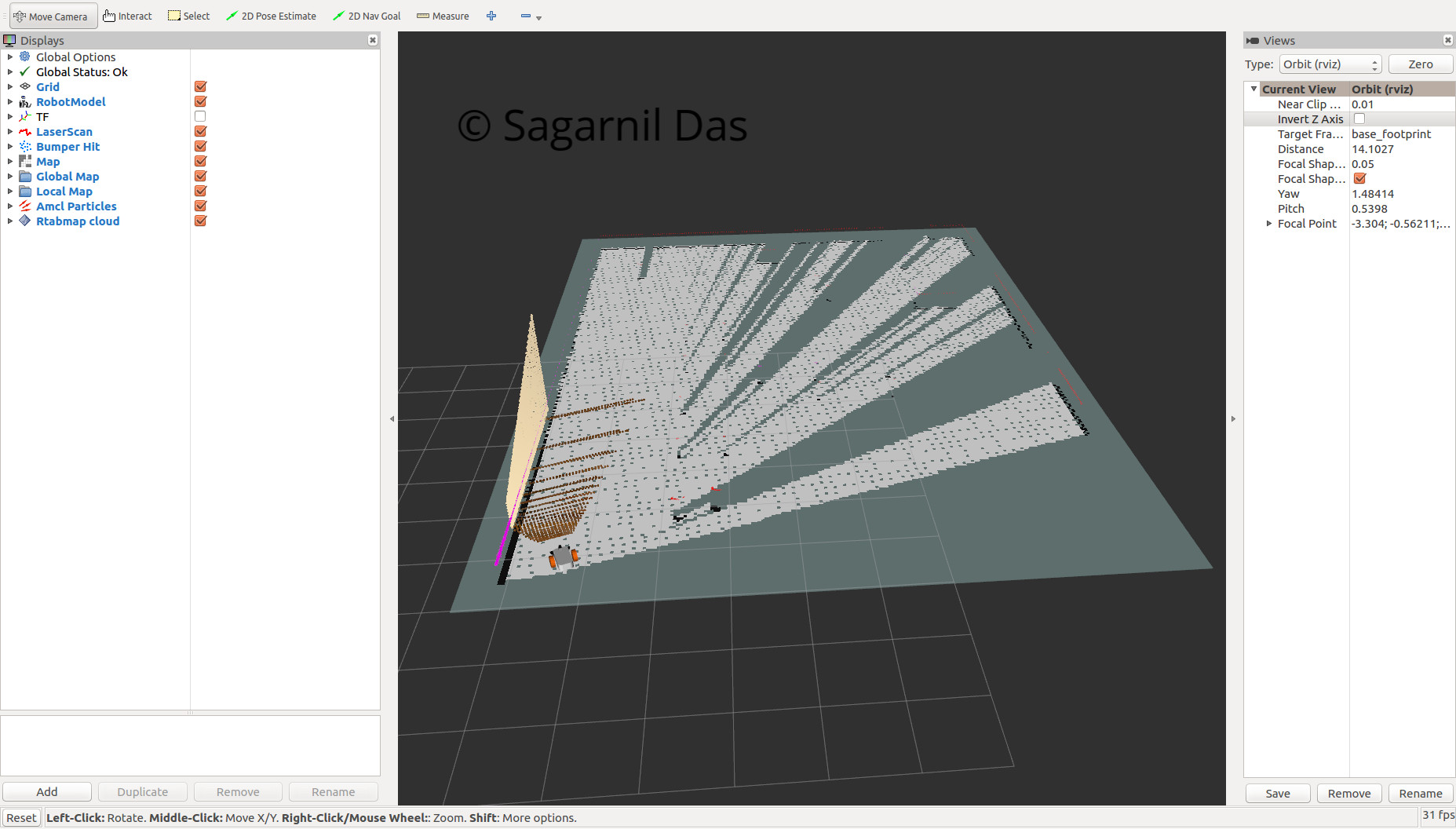}
      \caption{Starting point of the robot as seen in RViz}
      \label{fig:rviz_start}
\end{figure}

During the mapping of the environment, the mapping data is saved in the rtabmap.db database. The localization.launch file can be started in order to localize the robot during the run.

\section{Results}
The mapping was done by the robot controlled by the teleop keyboard. In order to be able to have more than 3 loop closure detection, which was the project's benchmark, the robot was navigated through the full environment of both the worlds so that it could collect more images.

\subsection{Kitchen-Dining World}

The mapping run in the provided world ended with 66 global loop closures. This file has a size of 315 MB and is named as rtabmap\_kitchen\_dining.db. Figure 9 shows the robot's trajectory as well as the 2d occupancy grid map of the kitchen\_dining world.

\begin{figure}[thpb]
      \centering
      \includegraphics[width=\linewidth]{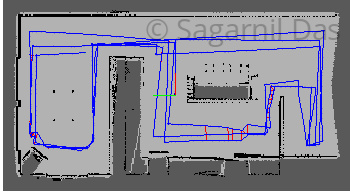}
      \caption{Robot's trajectory and 2d occupancy grid map of the Kitchen-Dining world}
      \label{fig:2d_world_udacity_world}
\end{figure}

At the end of the multiple passes, a well structured 3d point cloud map was created by using the \textit{Export 3d Clouds} functionality. Figure 10 depicts the reconstructed point cloud data. It can be seen that most features in the world like the chairs and tables are reconstructed properly and are distinctive. Figure 11 shows the RViz result of the same world after the end of the mapping task.

\begin{figure}[thpb]
      \centering
      \includegraphics[width=\linewidth]{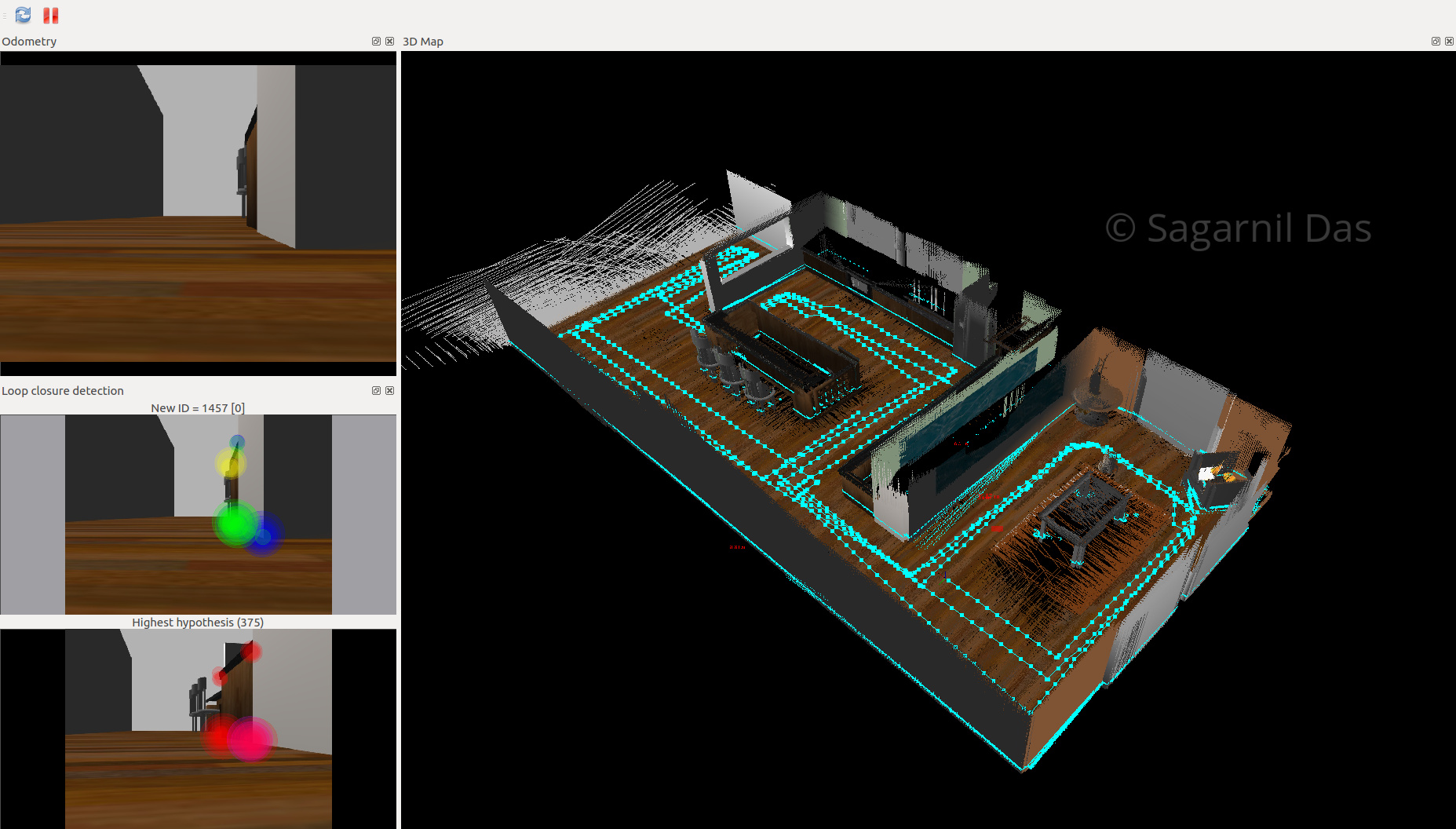}
      \caption{Reconstructed Point cloud data in RTAB-Map viewer of the Kitchen-Dining world}
      \label{fig:2d_world_udacity_world}
\end{figure}

\begin{figure}[thpb]
      \centering
      \includegraphics[width=\linewidth]{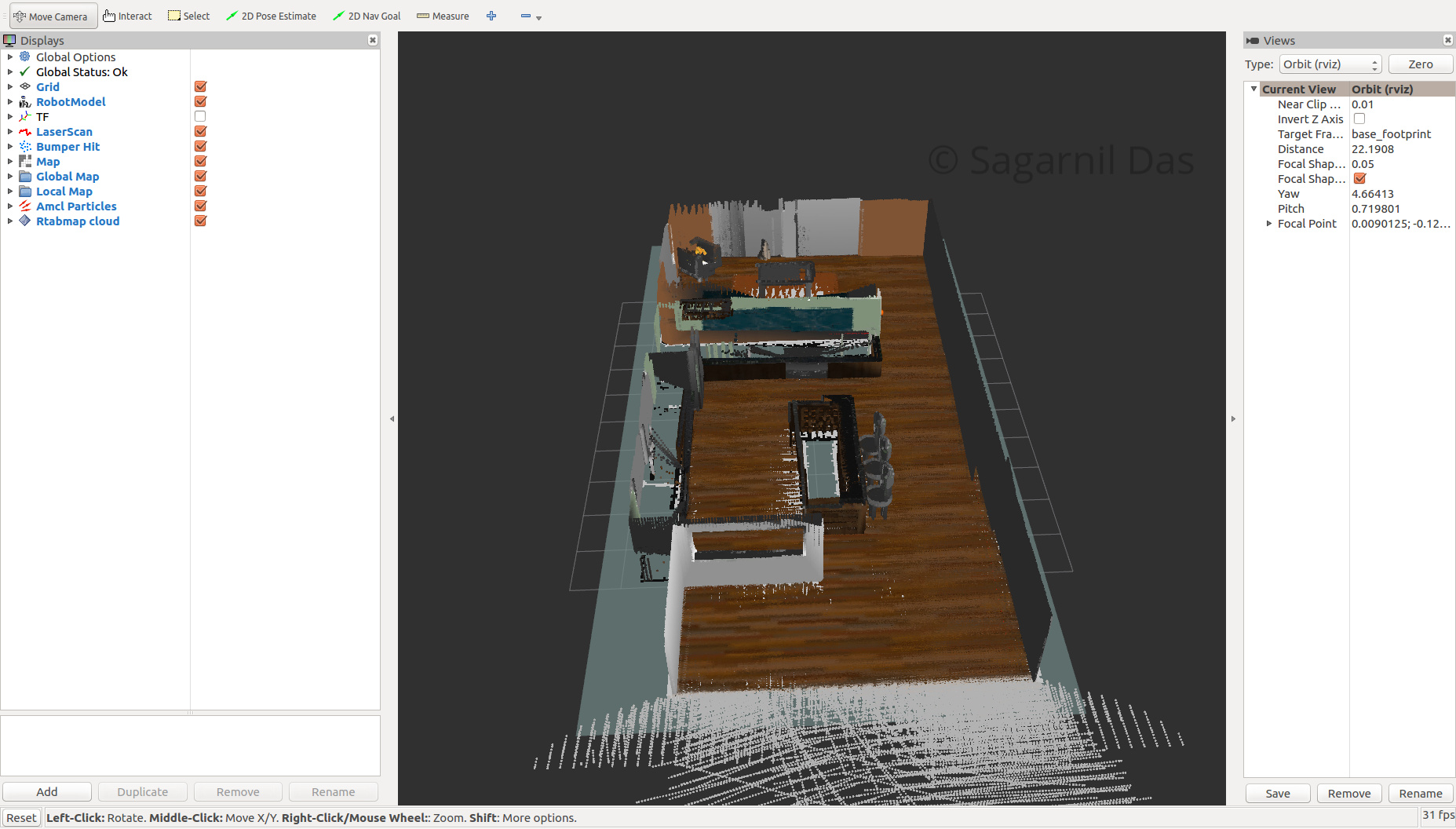}
      \caption{Rviz view of the Kitchen-dining world}
      \label{fig:rviz_udacity_world}
\end{figure}

At the end of the map, the loop closures can be seen in the rtabmap\_kitchen\_dining.db. Figure 12 shows one of them.

\begin{figure}[thpb]
      \centering
      \includegraphics[width=\linewidth]{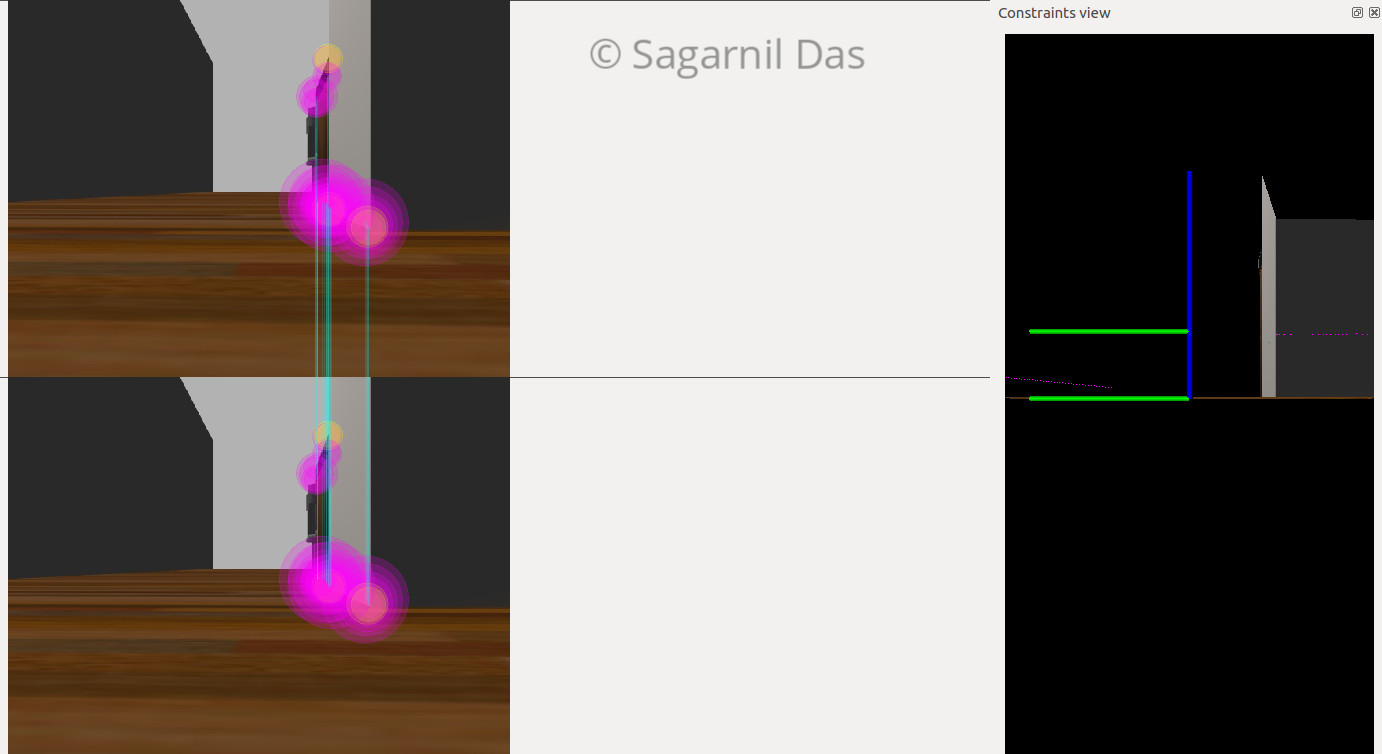}
      \caption{Loop Closure detection}
      \label{fig:loop_closure}
\end{figure}

\subsection{Sagar-cafe World} 
In the custom made world - \textbf{sagar\_cafe.world}, the robot performed well. As the robot is very short, some of the taller objects like the people, trees are not fully mapped. The kitchen also could not be traveled by the robot. Figure 13 and 14 depicts the RTAB-map view and the RViz view of this world respectively at the end of the mapping.

\begin{figure}[thpb]
      \centering
      \includegraphics[width=\linewidth]{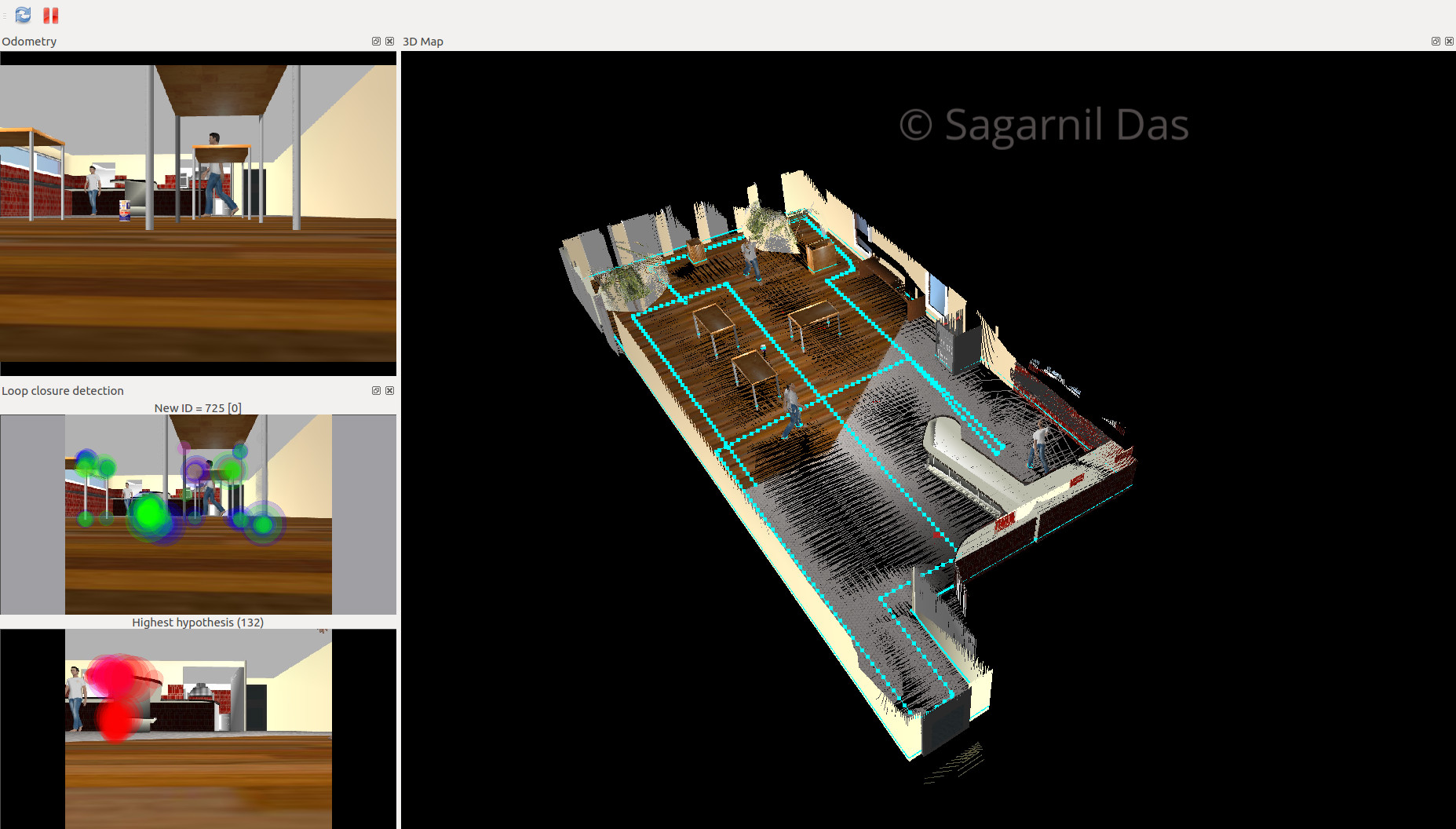}
      \caption{Reconstructed Point cloud data in RTAB-Map viewer in the sagar\_cafe world}
      \label{fig:rtabmap_sagar_cafe_world}
\end{figure}

\begin{figure}[thpb]
      \centering
      \includegraphics[width=\linewidth]{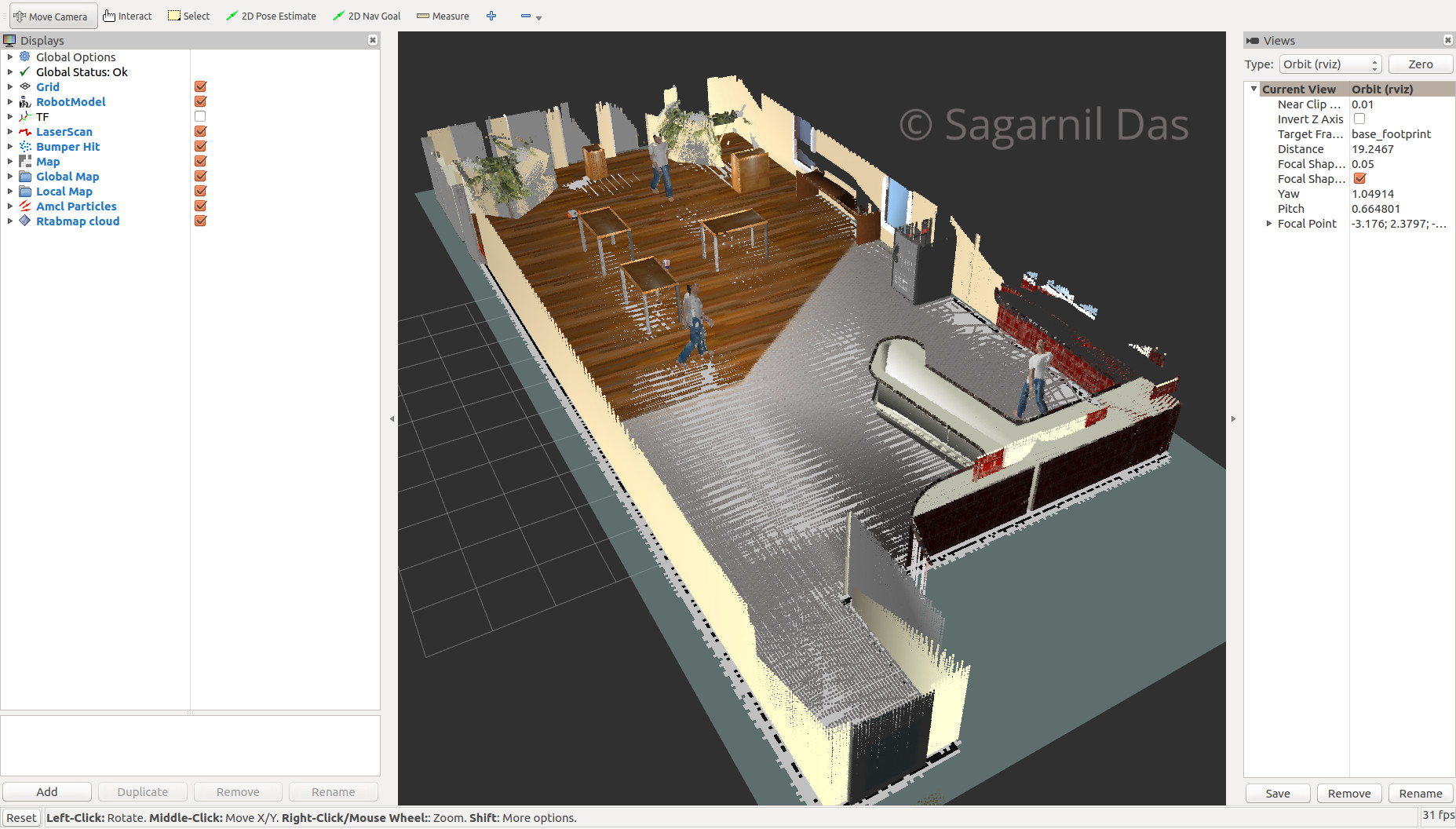}
      \caption{RViz view of the sagar\_cafe world}
      \label{fig:rviz_sagar_cafe_world}
\end{figure}

\section{Discussion}

In both the environments, successful mapping was performed in order to identify the ground truth and the distinctive features of the environment like, walls, tables, chairs, trees, people etc. In the sagar\_cafe world, the robot could not enter the kitchen area, which can be seen in right lower corner of Figure 13 and Figure 14. One of the possible explanations might be that there is a height difference of the floor between the living room and the kitchen. Another possible explanation might be that there is a transparent door separating those two rooms and hence the robot couldn't pass through. 

The generated 2d and 3d maps can be improved by doing more mapping runs which cover the environment in a more complete manner and by optimizing the loop closure detection further.

\section{Conclusion / Future work}
An interesting future work would be to explore the RTAB-Map package's visualization section in more details. The obstacle detection feature can be deployed in order to extracts obstacles and the ground from your point cloud. With this information in hand, these obstacles can be avoided when executing a desired path. Another potential area would be Wifi signal strength mapping. This feature allows the user to visualize the strength of your robot’s WiFi connection. This can be important in order to determine where the robot may lose its signal, therefore dictating it to avoid certain areas. The situation can also be remedied with larger antennas. 

\section{End sections}

\subsection{Acknowledgements}

The author sincerely thanks Udacity for their support and critiques throughout the project. The kitchen-dining world was provided by them.

\bibliography{bibfile}
\bibliographystyle{ieeetr}

\end{document}